\documentclass[final]{opt2026}

\usepackage{booktabs}       
\usepackage{microtype}      

\usepackage{titlesec}
\titlespacing*{\section}{0pt}{5pt plus 1pt minus 1pt}{2pt plus 1pt minus 1pt}
\titlespacing*{\subsection}{0pt}{4pt plus 1pt minus 1pt}{1pt plus 1pt minus 1pt}
\titlespacing*{\paragraph}{0pt}{3pt plus 1pt minus 1pt}{0.5em}

\title[Physical Muon]{Physical Muon: Orthogonalization as an Equilibrium Computation}

\optauthor{%
\Name{Yuren Hao} \Email{yurenh2@illinois.com}\\
\addr University of Illinois at Urbana-Champaign}

\begin{document}

\maketitle

\begin{abstract}
Physical neural networks and analog in-memory computing could reduce the energy cost of neural network training. Realizing this potential, however, requires optimizers that combine effective learning with physical implementability. SGD fits local analog updates but struggles on transformers, while Adam family is unstable against analog bias. Muon offers strong training performance, but its Newton--Schulz orthogonalization relies on dense matrix-matrix products. To address this obstacle, we introduce Physical Muon, which computes the orthogonalization as the equilibrium of a continuous-time flow. Random probes approximate the flow using matrix-vector products, reciprocal reads, and local rank-1 writes. To test whether this replacement preserves training performance, we evaluate it on a 10.95M-parameter transformer. The dense flow's mean validation cross-entropy is 0.0085 above Newton--Schulz across nine seeds per method; the probe implementation is 0.0188 above the control across two seeds. Circuit simulations further reproduce the flow dynamics and yield comparable training behavior.
\end{abstract}

\section{Introduction}
\label{sec:intro}

Physical neural networks and analog in-memory computing offer a route to reducing the energy cost of neural computation \citep{momeni2025training, wright2022deep, sebastian2020memory}. For example, resistive arrays store weights as conductances and compute matrix-vector products by summing currents, avoiding the movement of individual weights to a digital processor \citep{rpu, ambrogio2018equivalent}. Extending these advantages to training, however, requires an optimizer that both learns effectively and operates within the constraints of the physical hardware.

SGD's local updates fit resistive arrays \citep{rpu, tikitaka}, but its performance on transformers can fall substantially behind Adam \citep{zhang2024transformers}. Adam's adaptive updates address this training-quality gap by maintaining first and second moments and normalizing each parameter separately \citep{kingma2015adam, adamw}. However, systematic analog bias \citep{dillavou2025imperfection} can destabilize Adam's adaptive updates. Physical training therefore requires an optimizer that combines effective learning with robustness to analog bias.

These requirements motivate Muon, which orthogonalizes its momentum matrix and has demonstrated better training compute efficiency than AdamW in language models \citep{muon, muonscale}. Its standard Newton--Schulz (NS) implementation, however, requires dense matrix-matrix products. The target resistive arrays lack this native operation and must decompose a dense product into repeated matrix-vector reads \citep{rpu, sebastian2020memory}. To overcome this obstacle, we introduce Physical Muon, which computes the orthogonalization through a continuous-time relaxation directly expressible with array reads and local rank-1 writes.

To evaluate this formulation, we compare dense numerical integration and a random-probe implementation with NS on a 10.95M-parameter transformer. Both obtain validation loss close to NS, and ablations separately test output reuse and orthogonalization accuracy. These training experiments use digital simulation and standard backpropagation. We then assess physical implementability through device-error tests and circuit simulations, including a small synthetic task with the circuit in the training loop.

\paragraph{Related work.}
Recent work accelerates digital orthogonalization with improved polynomials and hardware-aware reformulations \citep{polarexpress, gramns}, or uses temporal information \citep{cachemuon}. Continuous-time matrix flows are classical \citep{oja, brockett}, and recent work studies Muon's parameter trajectory across steps \citep{wasserflow, hamflow, muonspectral}. Our contribution is an analog execution model for Muon's per-step orthogonalization, validated through training and circuit simulation. Physical Muon complements methods that obtain gradients through physical relaxation \citep{ep, clln1, clln3}.

\section{Physical Muon}
\label{sec:physical-muon}

At each optimizer step, Muon accumulates the gradient $g_t$ into a momentum buffer $m_t = \mu m_{t-1} + g_t$ and forms the Nesterov momentum $u_t = g_t + \mu m_t$. It orthogonalizes $u_t$ to obtain $O_t$, then updates the weight matrix as $W_{t+1} = (1 - \mathrm{lr}\,\lambda)W_t - \mathrm{lr}\,sO_t$, with learning rate $\mathrm{lr}$, weight decay $\lambda$, and shape-dependent scale $s$ \citep{muon, muonscale}. For full-rank $u_t = U\Sigma V^T$, the target is the polar factor $UV^T$, which preserves the singular vectors and maps the singular values to one \citep{higham}. This is the steepest descent direction under a spectral-norm constraint \citep{bernstein2024optimizer}; standard Muon approximates it with NS iterations.

\paragraph{Equilibrium formulation.}
To compute this transformation through relaxation, we hold the current momentum fixed and evolve an auxiliary matrix $X$ from zero:
\begin{equation}
\dot{X} = \hat{M} - XX^T\hat{M},
\qquad X(0) = 0,
\qquad \hat{M} = u_t/\alpha(u_t).
\label{eq:flow}
\end{equation}
Here $\alpha(u_t)>0$ is a scalar normalizer. Starting from zero keeps $X$ in the singular-vector basis of $\hat{M}$, so each singular value evolves independently:
\begin{equation}
\dot{d}_i = \hat{s}_i(1 - d_i^2),
\qquad d_i(t) = \tanh(\hat{s}_i t),
\label{eq:modes}
\end{equation}
where $\hat{s}_i$ is a singular value of $\hat{M}$. Every nonzero mode approaches one, giving the polar factor at equilibrium; zero modes remain zero, as in odd-polynomial NS iterations. At finite time, smaller modes remain attenuated. Section~\ref{sec:evaluation} evaluates the resulting effect on training by varying the integration time and comparing the flow with NS5 and exact orthogonalization.

An analog implementation also needs a practical way to normalize the momentum. The Frobenius norm $\|u_t\|_F$, computed from a sum of squared entries, offers an alternative to computing the largest singular value $\sigma_{\max}$. For any positive scalar normalizer, \eqref{eq:modes} preserves the limiting polar factor while rescaling the convergence rates. Relative to our $\sigma_{\max}$-normalized reference, Frobenius normalization slows the modes by $\|u_t\|_F/\sigma_{\max}$, whose measured median is 2.07 (Appendix~\ref{app:b}). We therefore double the solve budget in the normalization ablation of Section~\ref{sec:evaluation} to test this hardware-compatible alternative.

For numerical evaluation we use $T$ Euler iterations with step size $\eta$ and return $O_t=X_T$:
\begin{equation}
X_{k+1} = \Pi_{[-r,r]}\!\left[X_k + \eta\bigl(\hat{M} - X_k X_k^T \hat{M}\bigr)\right],
\label{eq:euler}
\end{equation}
where the element-wise clip models circuit supply rails and is inactive in the reported runs. We reset $X$ at each optimizer step. Since $\tanh(1.8)\approx0.95$, modes above $\hat{s}^{*}=1.8/(\eta T)$ reach approximately 95\% of their limiting value in the continuous-time solution. Figure~\ref{fig:transfer} shows how this threshold shifts with the solve budget. The flow can also stop when its relative change falls below a threshold, offering a measurable tradeoff between iteration count and training loss.

\paragraph{Analog execution.}
Random probes express the dense increment in \eqref{eq:euler} through array reads and local writes. For a probe $v$ with independent $\pm1$ entries \citep{hutchinson1989stochastic}, we compute
\begin{equation}
p = \hat{M}v, \qquad q = X^Tp, \qquad r = Xq, \qquad \Delta X = \eta(p-r)v^T.
\label{eq:probe}
\end{equation}
Since $\mathbb{E}[vv^T]=I$, the expected write at fixed $X$ is $\eta(\hat{M}-XX^T\hat{M})$, the dense Euler increment. The transposed product reads the same $X$ array in reverse, using reciprocity; cell $(i,j)$ is updated from only the row signal $p_i-r_i$ and column signal $v_j$ \citep{rpu, tikitaka, ambrogio2018equivalent}. We average $K$ probes per iteration, giving a budget of $B=3KT$ array passes, counted separately for each probe channel.

\begin{figure}[htbp]
\centering
\includegraphics[width=\linewidth]{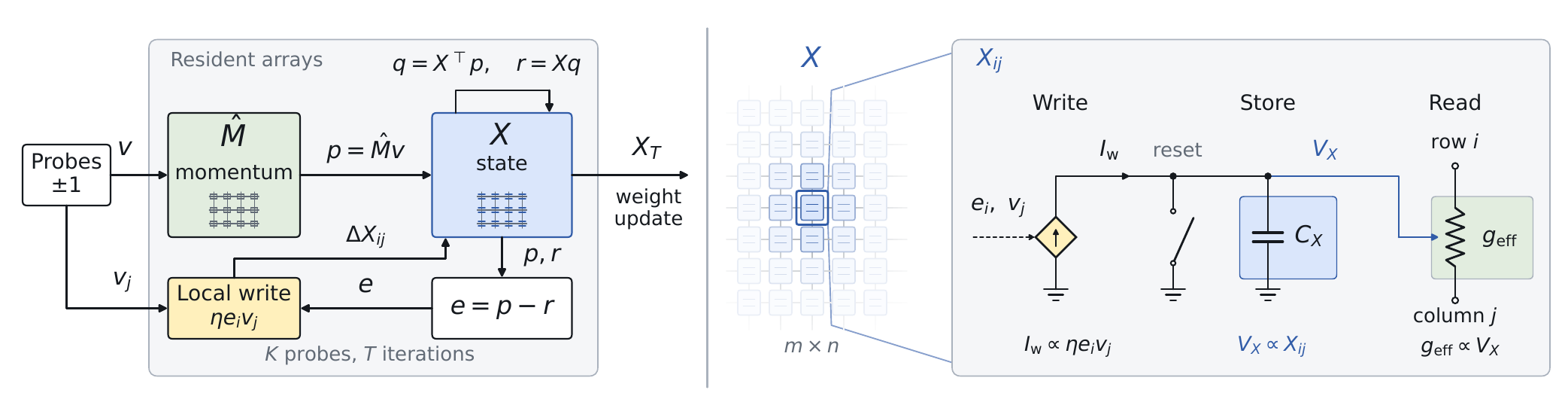}
\caption{Physical execution of \eqref{eq:probe}. Left: reciprocal reads form the row error for local rank-1 writes. Right: a portion of the $X$ array and an enlarged behavioral cell. A capacitor stores $X_{ij}$ as a voltage that controls its effective read conductance; signed write currents update the state. Section~\ref{sec:implementation} details circuit validation.}
\label{fig:arch}
\end{figure}

Figure~\ref{fig:arch} maps these operations to an array. Capacitors store the evolving state $X$; the signed row error charges each cell with a polarity set by its probe entry. The converged state supplies the weight update through the same read and rank-1 write primitives. We use zero initialization to obtain the singular-mode dynamics in \eqref{eq:modes}; Appendix~\ref{app:b} reports a preliminary warm-start comparison.

\section{Training Evaluation}
\label{sec:evaluation}

We test the replacement on a 12-layer, width-128 transformer (10.95M parameters), trained on FineWeb \citep{penedo2024fineweb} for 2,500 steps with batch size 24 and sequence length 256. We change only the orthogonalization, keeping standard backpropagation and the optimizer settings fixed: Muon uses momentum 0.95 and learning rate 0.016 on matrix parameters; AdamW updates the remaining parameters. We report best validation cross-entropy (CE), where lower is better. NS5 denotes five Newton--Schulz iterations. Full settings and per-seed results are in Appendix~\ref{app:c}.

\paragraph{Equilibrium-flow equivalence.}
We first test the relaxation itself using the \emph{dense flow}, which evaluates the full matrix update in \eqref{eq:euler} with $\eta=0.5$ and $T=400$. We assess practical equivalence with a CE margin of $\delta=0.0426$, the observed cost of reusing an NS5 output for one additional step (Section~\ref{sec:demands}). Across nine seeds per method, the dense flow passes the two one-sided tests (TOST) procedure \citep{schuirmann1987tost}. Its mean gap is 0.0085 and its one-sided 95\% upper bound is 0.0142, both within this margin (Table~\ref{tab:main_results}).

\begin{table}[htbp]
\centering
\caption{We perform an equivalence test using the two one-sided tests (TOST) procedure and find that NS5 and the dense flow have statistically equivalent validation CE within a margin of $\pm0.0426$ ($p=2.4\times10^{-8}$). CE is mean $\pm$ sample std across $n$ training seeds. $\Delta$CE is relative to NS5; the upper bound is its one-sided 95\% confidence bound.}
\label{tab:main_results}
\begin{tabular}{lrccc}
\toprule
Solver & $n$ & Validation CE & $\Delta$CE & Upper bound \\
\midrule
NS5 control & 9 & $5.0124\pm0.0081$ & --- & --- \\
Dense flow & 9 & $5.0209\pm0.0054$ & $+0.0085$ & $0.0142$ \\
\bottomrule
\end{tabular}
\end{table}

\paragraph{Hardware-compatible normalization.}
The normalization ablation changes the dense flow's normalizer from $\sigma_{\max}$ to $\|u_t\|_F$ and doubles $T$ to 800 to compensate for slower relaxation. Its CE is $5.0170\pm0.0083$ across four seeds, and it passes the same equivalence test with an upper bound of 0.0142 (Appendix~\ref{app:c}). Thus Frobenius normalization provides a practical alternative while preserving training quality within the chosen margin.

\paragraph{Array-operation approximation.}
We next test the \emph{probe flow}, which simulates the three reads and local writes in \eqref{eq:probe} inside transformer training. With $K=32$ probes along the shorter matrix dimension, $\eta=0.15$, and $T=417$, it uses about 40k passes per matrix per optimizer step. The two runs give a mean CE of 5.0313, or $+0.0188$ relative to NS5. This pilot tests training with the array-operation sequence; additional seeds would further characterize its variability. Circuit-in-the-loop training is evaluated in Section~\ref{sec:implementation}.

\subsection{Ablations and Generality}
\label{sec:demands}

\begin{figure}[t]
\centering
\includegraphics[width=0.49\linewidth]{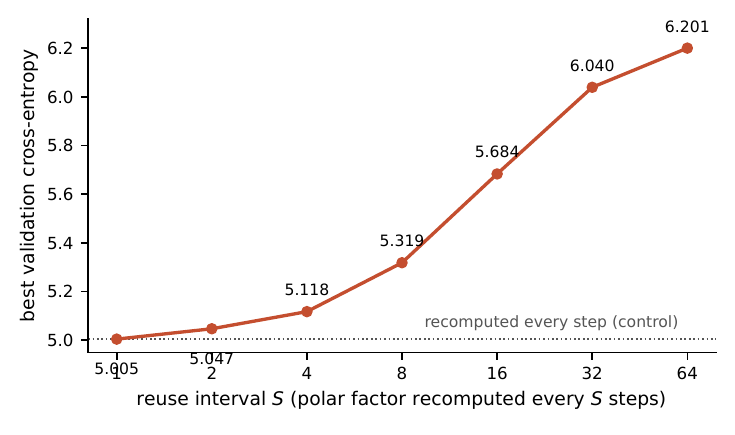}\hfill
\includegraphics[width=0.49\linewidth]{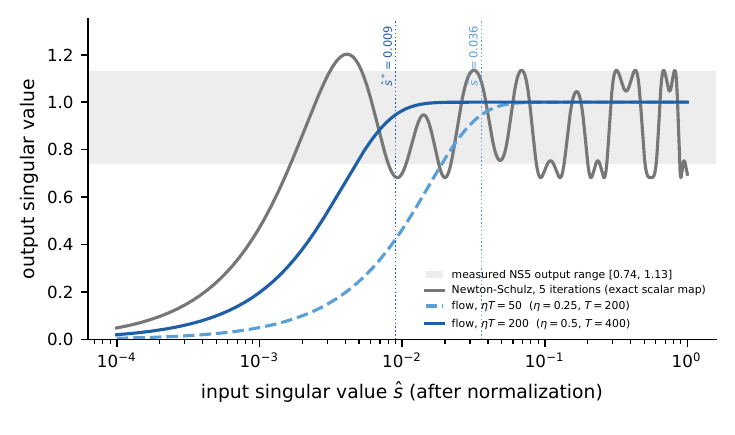}
\caption{Left: directly reusing NS5 outputs degrades training as the reuse interval grows (one seed). Right: singular-value responses of NS5 and the flow at two integration times; vertical lines mark the approximate 95\% thresholds. Shading covers the measured NS5 outputs.}
\label{fig:stale}
\label{fig:transfer}
\end{figure}

To interpret the replacement cost, we separately change output accuracy and freshness. Exact SVD gives a mean CE of 5.0119 across three seeds, close to NS5's 5.0124. SVD maps nonzero singular values exactly to one; the measured NS5 outputs span 0.74--1.13 (Figure~\ref{fig:transfer}, right). Thus exact singular-value scaling offers little observed benefit at this scale. The dense and probe experiments above directly assess our finite-time approximation. More accurate orthogonalization can improve training at larger scales \citep{polarexpress}.

Direct reuse behaves differently. Recomputing the NS5 output every $S$ steps and reusing it in between increasingly degrades training (Figure~\ref{fig:stale}, left). Even $S=2$ adds 0.0426 CE in the standard configuration. Saved momentum trajectories show rotation of the singular subspaces (Appendix~\ref{app:b}), explaining why an old output may be inaccurate for the current momentum. This ablation isolates direct reuse. Warm starts and history-based methods use past information to accelerate a fresh computation \citep{cachemuon, dion}.

The $S=2$ penalty supplies the reference margin in Table~\ref{tab:main_results}. The dense flow's upper confidence bound is about one third of this penalty, placing its replacement cost below the measured effect of reusing an output once. This margin is estimated from one seed and depends on the configuration; at batch size 6, reuse has little measured cost (Appendix~\ref{app:c}).

\paragraph{Solve duration.}
Short flow solves degrade training. At $\eta=0.25$, increasing $T$ from 50 to 200 reduces the CE gap from 0.136 to 0.033. This agrees with \eqref{eq:modes}: longer integration brings more small singular modes near one. An adaptive stop at relative change $3\times10^{-3}$ uses 96.7 iterations on average, versus 400 for the reference, with a gap of $+0.030$ on one seed. Thus the flow offers a tunable accuracy--cost tradeoff. Figure~\ref{fig:depth} and the full depth and stopping results are in Appendices~\ref{app:b}--\ref{app:c}.

\paragraph{Generality.}
Across widths corresponding to 10.95M--59M parameters, the dense-flow gap remains between 0.003 and 0.012, with one to nine seeds per setting. Reducing batch size from 24 to 3 tests noisier gradients; gaps range from 0.0078 to 0.0219 with at least three seeds per method. At a nearby learning rate, the dense flow's mean CE remains similar; extending training to 5,000 steps gives a gap of $+0.0010$ on one seed. These tests support a modest training cost across the studied settings, although the largest model and longer run each have only one seed (Appendix~\ref{app:c}).

\section{Physical Feasibility}
\label{sec:implementation}

The transformer experiments test the optimizer numerically. We now assess whether the array operations can be realized in a circuit, used inside training, and sustained under device errors and realistic operation budgets.

\paragraph{Circuit realization and training.}
An ngspice transient on an $8\times8$ momentum block executes the reciprocal reads and local writes in Figure~\ref{fig:arch}. An 8-bit differential resistor array stores the momentum, and 64 capacitors store $X$. Each iteration occupies $1\,\mu$s, including 40 ns read settling; the cold-start solve runs as one continuous 600-period transient. Its final state matches the numerical probe implementation at cosine 0.999983 and relative Frobenius distance 0.0059. Adding the modeled circuit non-idealities changes the cosine with NS5 by about 0.002. The $8\times8$ array uses behavioral elements; separate transistor-level read tests cover a $2\times2$ array (Appendix~\ref{app:e}).

\begin{figure}[htbp]
\centering
\includegraphics[width=\linewidth,trim={0 0 0 33bp},clip]{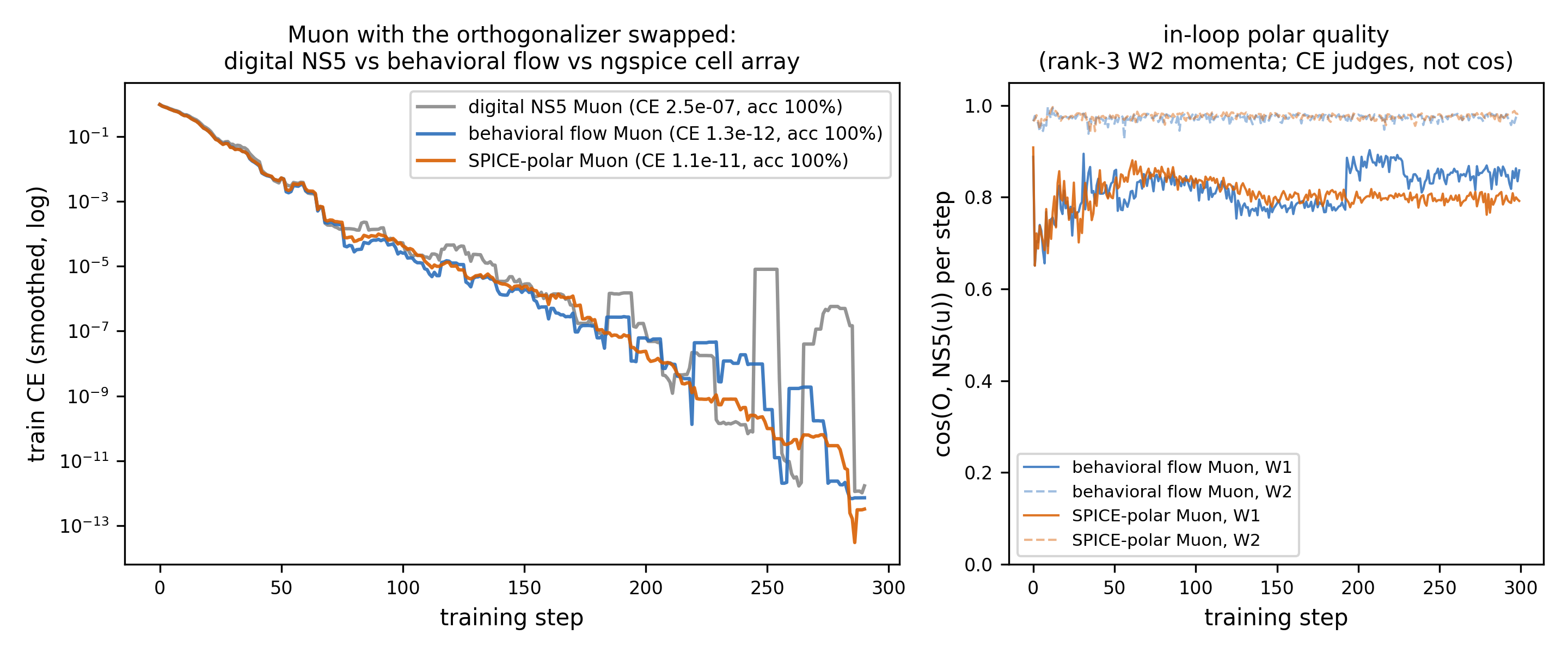}
\caption{Circuit-in-the-loop training on the four-class synthetic task. Left: training CE for the three Muon implementations defined above. Right: cosine similarity between their orthogonalized update directions and the NS5 reference for each of the two weight matrices. The momentum matrix for $W_2$ has rank three.}
\label{fig:toy}
\end{figure}

To test the circuit inside training, we use a four-class synthetic task (Figure~\ref{fig:toy}). The figure compares three implementations: \emph{digital NS5 Muon} uses five Newton--Schulz iterations; \emph{behavioral flow Muon} computes the probe flow with a numerical array model; and \emph{SPICE-polar Muon} executes the same flow through the ngspice circuit in the optimizer loop. All three reach CE below 0.01 in a similar number of updates and attain 100\% training accuracy. This small circuit-in-the-loop experiment complements the digital transformer evaluation.

\paragraph{Device errors.}
On three saved width-128 momentum shapes, the orthogonalized update from the probe flow reaches cosine similarity 0.9 with the NS5 update within 10k--40k array passes; the 40k budget used in transformer training covers all three shapes. At the selected budgets, 20\% persistent per-pass gain error changes this cosine by at most 0.011. At 10\% gain error, the cosine between perturbed and clean NS5 updates falls to 0.40--0.45, and PolarExpress diverges on two of the three matrices. A persistent offset of 1\% of signal rms and zero-mean write noise of 30\% change the probe output's cosine with NS5 by at most 0.0004 and 0.007, respectively (Appendix~\ref{app:d}). These tests quantify the directional robustness of orthogonalized updates on saved momentum matrices.

\paragraph{Energy and scaling.}
For the GPT-2 124M workload, projected orthogonalization energy is 0.45--1.04 J per optimizer step at 60k array passes per matrix, compared with an estimated 1.38--3.00 J for digital NS5. At 240k passes, the projection rises to 1.81--4.18 J, overlapping the digital range. The potential energy advantage therefore depends on the required solve budget.

These budgets extrapolate the measured 10k--40k passes on width-128 matrices linearly to rank 768. The energy ranges use published ReRAM and phase-change chip costs of 88.9 and 205 fJ per multiply-accumulate, respectively \citep{hung2021reram, legallo2023pcm}; the digital range spans peak-throughput and utilization-adjusted estimates. The corresponding break-even budgets are 104--517 passes per unit rank. These estimates assume resident arrays and exclude momentum loading, output readout, and shared buffer traffic. Appendix~\ref{app:d} gives the cost model, the original small-matrix budgets, and the latency assumptions.

\section{Discussion and Future Directions}
\label{sec:discussion}

To our knowledge, Physical Muon provides the first analog-compatible realization of Muon's orthogonalization, using reciprocal matrix-vector reads and local rank-1 writes. The dense flow meets the chosen practical-equivalence margin in transformer training, and circuit-in-the-loop simulations demonstrate these operations on a synthetic task. Open problems include scaling the probe budget to larger matrices, sustaining closed-loop training under measured device errors, and integrating the optimizer with physical gradient computation. These directions connect the present orthogonalization mechanism to end-to-end physical training.

\bibliography{references}


\appendix

\section{Optimizer Comparison}
\label{app:a}

Table~\ref{tab:zoo} reports a supplementary optimizer comparison. All arms train the same 12-layer, width-128 transformer for the same number of steps with backpropagation, with no weight decay; the learning rate of each method is swept until the optimum is bracketed. The reported quantity is the best validation CE.

\begin{table}[h]
\centering
\caption{Best validation CE for 11 optimizers under one protocol}
\label{tab:zoo}
\begin{tabular}{lc}
\toprule
Optimizer & CE \\
\midrule
Muon \citep{muon} & 4.9190 \\
OLion \citep{olion} & 4.9263 \\
OLion, 3 NS rounds & 4.9523 \\
OLion, 2 NS rounds & 5.0043 \\
AdamW \citep{adamw} & 5.0707 \\
OLion, 1 NS round & 5.1503 \\
Adafactor \citep{adafactor} & 5.2094 \\
Cautious Lion \citep{cautious} & 5.2163 \\
Lion \citep{lion} & 5.2242 \\
OLion, 0 NS rounds & 5.2604 \\
SGD with momentum & 5.3721 \\
\bottomrule
\end{tabular}
\end{table}

The OLion rows with 0 to 5 Newton-Schulz rounds form an ablation series; three rounds land within 0.026 of five.

\section{Rotation-Rate Measurements and Telemetry}
\label{app:b}

\paragraph{Rotation of the singular subspace.} The measurements use momentum matrices saved every step over 601 steps of a width-128 run, evaluated on three matrix shapes in a late-training window. For a stale interval of $k$ steps, the NS5 output of the matrix saved $k$ steps earlier is compared with the NS5 output of the current matrix. Table~\ref{tab:rotation} lists the cosine over the full matrix and the cosine after projection onto the current leading-16 singular subspace. Between 117 and 126 of the 128 singular modes carry singular values below half of the leading value.

\begin{table}[h]
\centering
\caption{Cosine between stale and fresh NS5 outputs}
\label{tab:rotation}
\begin{tabular}{lccc}
\toprule
Stale interval (steps) & 5 & 10 & 20 \\
\midrule
Full matrix & 0.60 & 0.39 & 0.19 \\
Leading-16 subspace & 0.89 & 0.78 & 0.52 \\
\bottomrule
\end{tabular}
\end{table}

\paragraph{Normalization.} On 174 saved momentum samples across three shapes, $\|u\|_F/\sigma_{\max}$ ranges from 1.02 to 3.19 (median 2.07), below the worst-case bound $\sqrt{\mathrm{rank}}\approx11.3$. Doubling the Frobenius-normalized solve to $T=800$ approximately compensates the median slowdown; its training result is reported in Section~\ref{sec:evaluation}. At $T=400$, one seed gives CE 5.0226; at $T=1600$, one seed gives 5.0161. A multi-seed budget sweep would quantify the tradeoff between solve cost and training quality.

\paragraph{Reset comparison.} On three saved late-training momentum matrices, a 50-iteration solve initialized from the previous step's state produces an update with cosine 0.85--0.86 against the fresh NS output; a 100-iteration solve from zero produces an update with cosine 0.94. These measurements use different budgets; an equal-cost comparison would isolate the effect of initialization. We use zero initialization for the singular-mode dynamics in \eqref{eq:modes}.

\paragraph{Stopping-criterion telemetry.} With the relative update sampled every 10 iterations and a cap of 400, halting at a relative update of $3\times10^{-3}$ requires a mean of 96.7 iterations per matrix (median 93.8, range 72.2--132.5) over 60 matrices; no matrix reaches the cap. Training with this rule gives CE 5.0419 on one seed, $+0.030$ above the NS control mean.

\begin{figure}[h]
\centering
\includegraphics[width=0.85\linewidth]{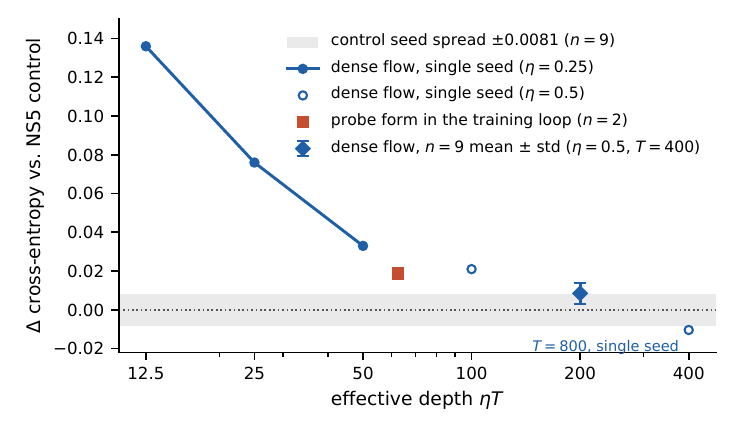}
\caption{Flow--control CE gap against integration time. Filled circles: $\eta=0.25$; open circles: $\eta=0.5$; diamond: $T=400$, nine-seed mean $\pm$ std; square: probe flow, $K=32$, $\eta=0.15$, $T=417$, two-seed mean. The grey band is the NS5 seed spread.}
\label{fig:depth}
\end{figure}

\paragraph{Depth series.} Figure~\ref{fig:depth} plots the flow--control gap against integration time $\eta T$ for the dense and probe implementations. Most depth settings have one seed; the diamond and square aggregate nine and two seeds, respectively.

\section{Training Protocol and Full Language-Model Results}
\label{app:c}

\paragraph{Protocol.} Unless varied explicitly, the language-model experiments use a 12-layer, width-128 transformer (10.95M parameters), trained with standard backpropagation on FineWeb \citep{penedo2024fineweb} for 2,500 steps at sequence length 256 and batch size 24, in fp32. Muon applies momentum 0.95 with Nesterov and learning rate $1.6\times10^{-2}$ to two-dimensional matrix parameters, with no weight decay. AdamW governs the remaining parameters at learning rate $10^{-3}$ and weight decay $10^{-4}$. A 250-step linear warmup precedes cosine decay to $0.1\times$. All CE values are the best validation CE. The optimizer comparison in Appendix~\ref{app:a} uses its separately described learning-rate sweeps and weight-decay settings.

\paragraph{Primary comparison, per seed.} Table~\ref{tab:seeds} lists the individual seeds behind the $n=9$ comparison in Table~\ref{tab:main_results}.

\begin{table}[h]
\centering
\caption{Best validation CE per seed, NS5 control and flow ($T=400$, $\eta=0.5$)}
\label{tab:seeds}
\small
\begin{tabular}{lccccccccc}
\toprule
Seed & 1 & 2 & 3 & 4 & 5 & 6 & 7 & 8 & 9 \\
\midrule
NS5 control & 5.0048 & 5.0084 & 5.0183 & 5.0080 & 5.0233 & 5.0248 & 5.0144 & 5.0035 & 5.0064 \\
Flow & 5.0251 & 5.0152 & 5.0217 & 5.0185 & 5.0178 & 5.0285 & 5.0124 & 5.0220 & 5.0269 \\
\bottomrule
\end{tabular}
\end{table}

\paragraph{Exact polar factor.} The SVD arm records CE values of 5.0092, 5.0120, and 5.0146 across three seeds (mean 5.0119), against the $5.0124 \pm 0.0081$ NS5 control.

\paragraph{Probe implementation.} The two training runs use $K=32$, $\eta=0.15$, and $T=417$, with probes along the shorter side of each matrix. Their CE values are 5.0281 and 5.0344, giving a mean of 5.03125, $+0.0188$ above the NS5 mean and $+0.0103$ above the dense flow. The budget is $3KT=40{,}032$ array passes per matrix per optimizer step.

\paragraph{Equivalence test.} With the reference margin $\delta=0.0426$ from the standard-configuration reuse test, TOST gives $p=2.4\times10^{-8}$ for the dense flow ($t=10.57$, Welch df 14). Its one-sided 95\% upper confidence bound on the CE difference is 0.0142; the test passes down to approximately this margin ($p=0.049$ at 0.0142). A two-sided test detects a small positive CE difference ($p=0.020$), which lies within the chosen equivalence margin. The four-seed Frobenius-normalized flow at $T=800$ also passes at $\delta=0.0426$ ($p=1.6\times10^{-4}$), with the same rounded upper bound of 0.0142.

\paragraph{Reuse penalty in other configurations.} The $S=2$ penalty that sets the equivalence margin in Section~\ref{sec:demands} is $+0.0426$ at width 128 (seed 1). At width 256 the paired $S=2$ penalty is $+0.0174$, above the 0.0142 bound achieved in Section~\ref{sec:evaluation}. At batch size 6 the $S=2$ penalty is $+0.0025$, inside the seed noise, so the freshness argument applies to the standard batch size.

\paragraph{Depth series.} Table~\ref{tab:depth} lists the integration-depth series from Section~\ref{sec:evaluation} together with the approximate mode threshold $\hat{s}^* = 1.8/(\eta T)$ of each configuration. An additional single-seed run at $T=800$, $\eta=0.5$ gives CE 5.0147, 0.0104 below its paired control.

\begin{table}[h]
\centering
\caption{Flow degradation against the NS5 control across integration depths (seed 1)}
\label{tab:depth}
\begin{tabular}{lcccc}
\toprule
$\eta$ & $T$ & $\eta T$ & $\hat{s}^*$ & $\Delta$CE \\
\midrule
0.25 & 50  & 12.5 & 0.144 & $+0.136$ \\
0.25 & 100 & 25   & 0.072 & $+0.076$ \\
0.25 & 200 & 50   & 0.036 & $+0.033$ \\
0.5  & 200 & 100  & 0.018 & $+0.021$ \\
0.5  & 400 & 200  & 0.009 & $+0.0085^{*}$ \\
\bottomrule
\end{tabular}

\smallskip
{\small $^{*}$ $n=9$ mean; the other rows are seed-1 measurements.}
\end{table}

\paragraph{Learning rate and training duration.} At learning rate $1.42\times10^{-2}$, the dense flow gives CE $5.0205\pm0.0119$ ($n=5$), compared with $5.0209\pm0.0054$ ($n=9$) at $1.6\times10^{-2}$. NS5 at $1.42\times10^{-2}$ or $1.8\times10^{-2}$ is approximately $+0.010$ worse than at $1.6\times10^{-2}$ in single-seed comparisons. Extending training to 5,000 steps reduces the flow--control gap to $+0.0010$ on one seed.

\paragraph{Gradient-noise series.} Table~\ref{tab:noise} lists the flow--control gap across batch sizes at the NS learning rate. Every arm contains at least three seeds; 17 of 21 paired seed differences are positive.

\begin{table}[h]
\centering
\caption{Flow--control gap across batch sizes}
\label{tab:noise}
\begin{tabular}{lccccc}
\toprule
Batch size & 24 & 12 & 6 & 4 & 3 \\
\midrule
$\Delta(\text{flow} - \text{ctl})$ & $+0.0085$ & $+0.0175$ & $+0.0078$ & $+0.0219$ & $+0.0149$ \\
\bottomrule
\end{tabular}
\end{table}

\paragraph{Width series.} Table~\ref{tab:width} lists the flow--control gap across model widths. The gap stays within $[0.003, 0.012]$ and shows no monotone direction.

\begin{table}[h]
\centering
\caption{Flow--control gap across widths at $T=400$}
\label{tab:width}
\begin{tabular}{lcccc}
\toprule
Width & 128 (10.95M) & 192 & 256 (26.4M) & 384 (59M) \\
\midrule
$\Delta$ & $+0.0085$ & $+0.0114$ & $+0.0034$ & $+0.0112$ \\
Seeds & 9 & 3 & 2 & 1 \\
\bottomrule
\end{tabular}
\end{table}

\section{Probe Budget Frontier, Device Tolerance, and Energy Accounting}
\label{app:d}

\paragraph{Frontier.} Three saved momentum matrices from a width-128 run (a $128 \times 128$ attention matrix and the $384 \times 128$ and $128 \times 384$ feed-forward matrices) serve as test inputs. For each budget $B$ of array passes ($B = 3KT$ for $K$ parallel probe channels and $T$ iterations), the grid over $K \in \{8, 16, 32, 64\}$ and $\eta \in \{0.15, 0.3, 0.5\}$ is searched with three probe seeds, and Table~\ref{tab:frontier} reports the best cosine between the probe-flow output and the digital NS5 output. The step size $\eta = 0.15$ is the best choice at every entry. The probe-form code path reproduces the dense flow bitwise when probes are replaced by the identity. For reference, the dense-flow outputs at $T=400$ have cosines 0.9833, 0.9854, and 0.9861 with the NS5 outputs on these three shapes, respectively. Cosines restricted to the leading-16 singular subspace range from 0.96 to 0.99 over the probe budgets.

\begin{table}[h]
\centering
\caption{Best cosine against NS5 (best $K$ in parentheses) across array-pass budgets; entries above the 0.9 screen in bold}
\label{tab:frontier}
\footnotesize
\setlength{\tabcolsep}{4pt}
\begin{tabular}{lcccccc}
\toprule
Budget $B$ & 2k & 5k & 10k & 20k & 40k & 80k \\
\midrule
$128 \times 128$ attention & 0.708 (16) & 0.802 (16) & 0.850 (16) & 0.888 (16) & \textbf{0.919} (32) & \textbf{0.945} (32) \\
$384 \times 128$ & 0.806 (8) & 0.896 (8) & \textbf{0.930} (16) & \textbf{0.951} (16) & \textbf{0.965} (32) & \textbf{0.973} (64) \\
$128 \times 384$ & 0.677 (16) & 0.803 (16) & 0.873 (32) & \textbf{0.912} (32) & \textbf{0.946} (64) & \textbf{0.953} (64) \\
\bottomrule
\end{tabular}
\end{table}

\paragraph{Device tolerance.} Each defect class is injected into the probe form at the frontier configuration of each matrix (three probe seeds), and Table~\ref{tab:tolerance} reports the change in the screen value. Gain errors are persistent per-pass scalars $1 + g\,\xi$ with $\xi$ drawn once; offsets are persistent per-pass vectors scaled to the signal rms; write noise is fresh zero-mean noise on each rank-1 write.

\begin{table}[h]
\centering
\caption{Change in screen value (cosine against NS5) under injected defects}
\label{tab:tolerance}
\small
\begin{tabular}{lccccccc}
\toprule
 & \multicolumn{3}{c}{Persistent gain} & \multicolumn{2}{c}{Persistent offset} & \multicolumn{2}{c}{Write noise} \\
\cmidrule(lr){2-4}\cmidrule(lr){5-6}\cmidrule(lr){7-8}
 & 5\% & 10\% & 20\% & 1\% & 5\% & 10\% & 30\% \\
\midrule
$128 \times 128$ attention & $+0.0004$ & $+0.0008$ & $+0.0013$ & $-0.0004$ & $-0.0129$ & $-0.0006$ & $-0.0052$ \\
$384 \times 128$ & $+0.0002$ & $+0.0003$ & $+0.0001$ & $-0.0001$ & $-0.0014$ & $-0.0002$ & $-0.0017$ \\
$128 \times 384$ & $+0.0005$ & $-0.0008$ & $-0.0106$ & $-0.0002$ & $-0.0073$ & $-0.0008$ & $-0.0069$ \\
\bottomrule
\end{tabular}
\end{table}

Table~\ref{tab:methodcross} compares three orthogonalizers under the same persistent gain error, with the metric defined as the cosine between the defected output and the clean output of the same method; the entries give the range over the three matrices. Uniform output scaling leaves this cosine unchanged. Separate read gains can change the flow's equilibrium amplitude as well as its timescale; the reported cosine measures directional agreement.

\begin{table}[h]
\centering
\caption{Cosine between defected and clean output under persistent per-pass gain error}
\label{tab:methodcross}
\small
\begin{tabular}{lccc}
\toprule
Method & 5\% & 10\% & 20\% \\
\midrule
NS5 & 0.913--0.940 & 0.397--0.447 & 0.366--0.424 \\
PolarExpress, 5 rounds & 0.366--0.429 & 2 of 3 diverged; 0.389 & 3 of 3 diverged \\
Flow (probe form) & 0.9986--0.9999 & 0.9945--0.9998 & 0.9763--0.9990 \\
\bottomrule
\end{tabular}
\end{table}

\paragraph{Energy model.} We use an operation-count estimate, $E(B)=B\,\Sigma_{rc}\,e$, with per-MAC energy $e$. This extrapolation from published silicon measurements is analogous to the component-based power estimate of \citet{wang2026analog}. The workload unit is GPT-2 124M: 48 momentum matrices containing $\Sigma_{rc}=84.9\times10^6$ cells. A fabricated ReRAM chip \citep{hung2021reram} provides a lower anchor of 10.3 fJ per multiply-accumulate (MAC) for the binary-input probe pass and 128 fJ for the other two passes, averaging 88.9 fJ. A fabricated phase-change chip \citep{legallo2023pcm} provides a higher anchor of 205 fJ per MAC. These published chip measurements include per-pass conversion circuitry and serve as cost anchors for the proposed loop. The digital model assumes $9.784\times10^{11}$ MACs for NS5 on an H100, giving 1.38 J at datasheet peak \citep{h100datasheet} or 3.00 J using the 46.2\% utilization reported for large-scale training \citep{palm} as a proxy for NS5 utilization.

\paragraph{Rank scaling and break-even.} Starting from zero with inactive rails, $KT$ rank-1 writes give $\mathrm{rank}(X)\le KT=B/3$ under the counted write model. The measured width-128 budgets span 78--312 passes per unit rank. Extrapolating these budgets linearly to rank 768 gives 60k--240k passes; validating this projection requires width-768 probe training. Table~\ref{tab:energy} applies both the measured small-matrix budgets and the rank-scaled budgets to the same GPT-2 workload. Break-even lies at 239--517 passes per unit rank for the lower anchor and 104--224 for the higher anchor, depending on digital utilization. At 240k passes the projected range overlaps the digital estimate.

\paragraph{Latency and excluded costs.} A simulated iteration takes 1.0 $\mu$s. Assuming all matrices are resident and the $K$ probe channels operate in parallel at this period, the projected solve time for $T=417$ is 0.42 ms, compared with the 2 ms digital reference used in the accounting. A system-level evaluation would test these residency and parallel-execution assumptions. The energy totals exclude loading $\hat M$ (estimated at 0.85 mJ per step using a measured DAC \citep{olieman2015dac}), reading $X$ out once per step, and 1.7 GB of buffer traffic shared by both implementations.

\begin{table}[h]
\centering
\caption{Estimated orthogonalization energy per step for the GPT-2 124M workload. The 10k--40k budgets are measured on width-128 matrices; 60k--240k are rank-scaled projections. Pitch and stress denote the lower and higher device energy anchors.}
\label{tab:energy}
\small
\begin{tabular}{lcccc}
\toprule
Budget per matrix & $E$ (pitch) & $E$ (stress) & Digital / pitch & Digital / stress \\
\midrule
$B = 10$k (measured, width 128) & 0.076 J & 0.174 J & 18.3--39.7$\times$ & 8.0--17.2$\times$ \\
$B = 20$k (measured, width 128) & 0.151 J & 0.348 J & 9.2--19.8$\times$ & 4.0--8.6$\times$ \\
$B = 40$k (measured, width 128) & 0.302 J & 0.696 J & 4.6--9.9$\times$ & 2.0--4.3$\times$ \\
$B = 60$k (rank-scaled, $r = 768$) & 0.453 J & 1.044 J & 3.1--6.6$\times$ & 1.3--2.9$\times$ \\
$B = 240$k (rank-scaled, $r = 768$) & 1.812 J & 4.177 J & 0.8--1.7$\times$ & 0.3--0.7$\times$ \\
\bottomrule
\end{tabular}
\end{table}

\section{Circuit-Level Validation}
\label{app:e}

\paragraph{Simulation ladder.} Three levels connect the mathematical flow to a circuit: the dense Euler iteration of \eqref{eq:euler}, the probe form of \eqref{eq:probe} with a behavioral array model, and an ngspice transient of the array. Table~\ref{tab:ladder} lists the agreement at each interface.

\begin{table}[h]
\centering
\caption{Agreement across the simulation ladder}
\label{tab:ladder}
\footnotesize
\setlength{\tabcolsep}{4pt}
\begin{tabular}{lll}
\toprule
Interface & Quantity & Value \\
\midrule
Probe form vs.\ dense flow & code path with identity probes & bitwise equal \\
Quantized vs.\ continuous block & cosine of NS5 outputs & 1.0000 \\
Circuit vs.\ probe form & cosine of converged states & 0.999983 \\
 & relative Frobenius distance & 0.0059 \\
 & maximum element-wise distance & 0.0045 \\
Circuit vs.\ NS5 & screen value, clean & 0.9627 (behavioral 0.9624) \\
 & screen value, with non-idealities & 0.9606 \\
 & dense flow at $T = 400$ on the same block & 0.9834 \\
Closed loop (synthetic task) & steps to CE $< 0.01$, NS5 / behavioral / circuit & 43 / 42 / 42 \\
 & final CE, all three & 0.0000 \\
\bottomrule
\end{tabular}
\end{table}

\begin{figure}[h]
\centering
\includegraphics[width=\linewidth]{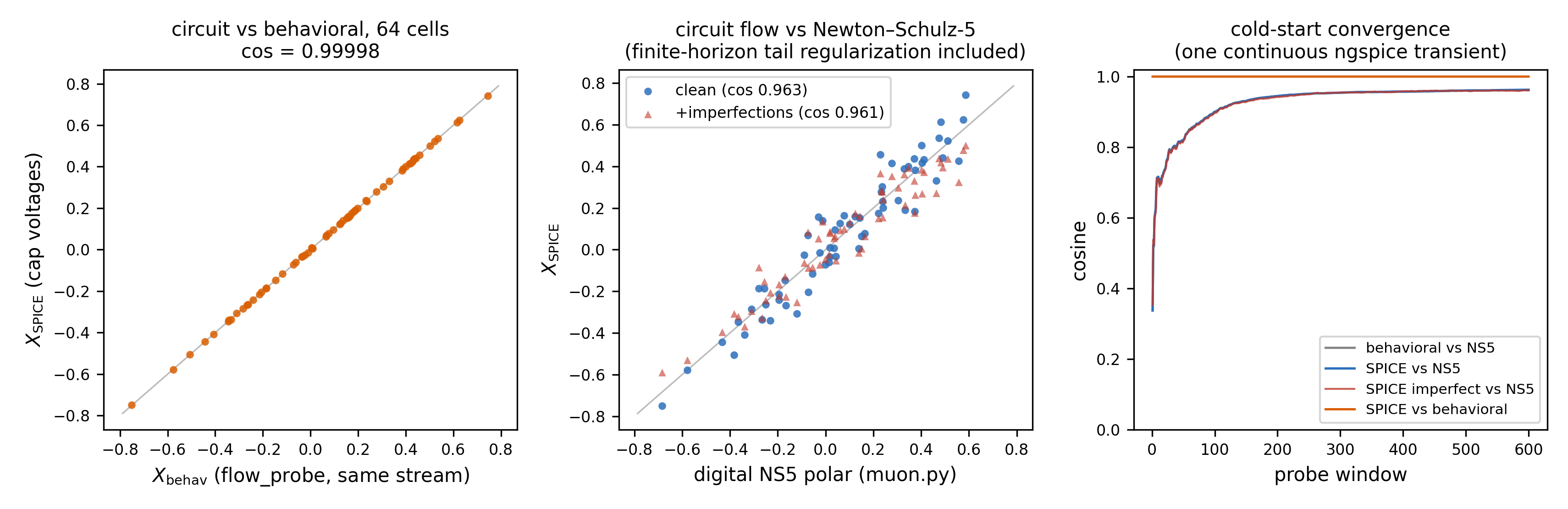}
\caption{Circuit transient on the $8 \times 8$ block. Left: the 64 capacitor voltages against the probe-form state (cosine 0.999983). Middle: screen values of the circuit output against NS5, clean and with non-idealities. Right: convergence over the 600-period transient.}
\label{fig:cellB}
\end{figure}

\paragraph{Array model.} The momentum array is a differential resistor network with 8-bit conductance codes and 0.4\% mismatch per resistor; its currents sum into a 0 V virtual ground, and it is written digitally once per optimizer step. The state $X$ is stored on 64 capacitors (one per entry of the $8 \times 8$ block). The passes $q = X^T p$ and $r = Xq$ read the same stored values, which is the reciprocity the transposed pass relies on. Each pass settles through a 40 ns RC time; the write gate is open for 0.60--0.95 $\mu$s of each 1.0 $\mu$s period, and the complete cold-start solve runs as one continuous transient of 600 periods (9.0 s of ngspice time clean, 12.5 s with non-idealities, single core). The test block is the $8 \times 8$ leading sub-block of a saved width-128 attention momentum matrix, with singular values from $3.65\times10^{-3}$ to $1.2\times10^{-4}$. The element-wise rail at 1.2 stays inactive ($\max|X| = 0.749$ clean, 0.592 with non-idealities). Figure~\ref{fig:arch} (right) shows one $X$ cell; Figure~\ref{fig:cellB} shows the capacitor voltages against the probe-form state, the cosine values against NS5, and the convergence trajectory.

\begin{figure}[h]
\centering
\includegraphics[width=0.48\linewidth]{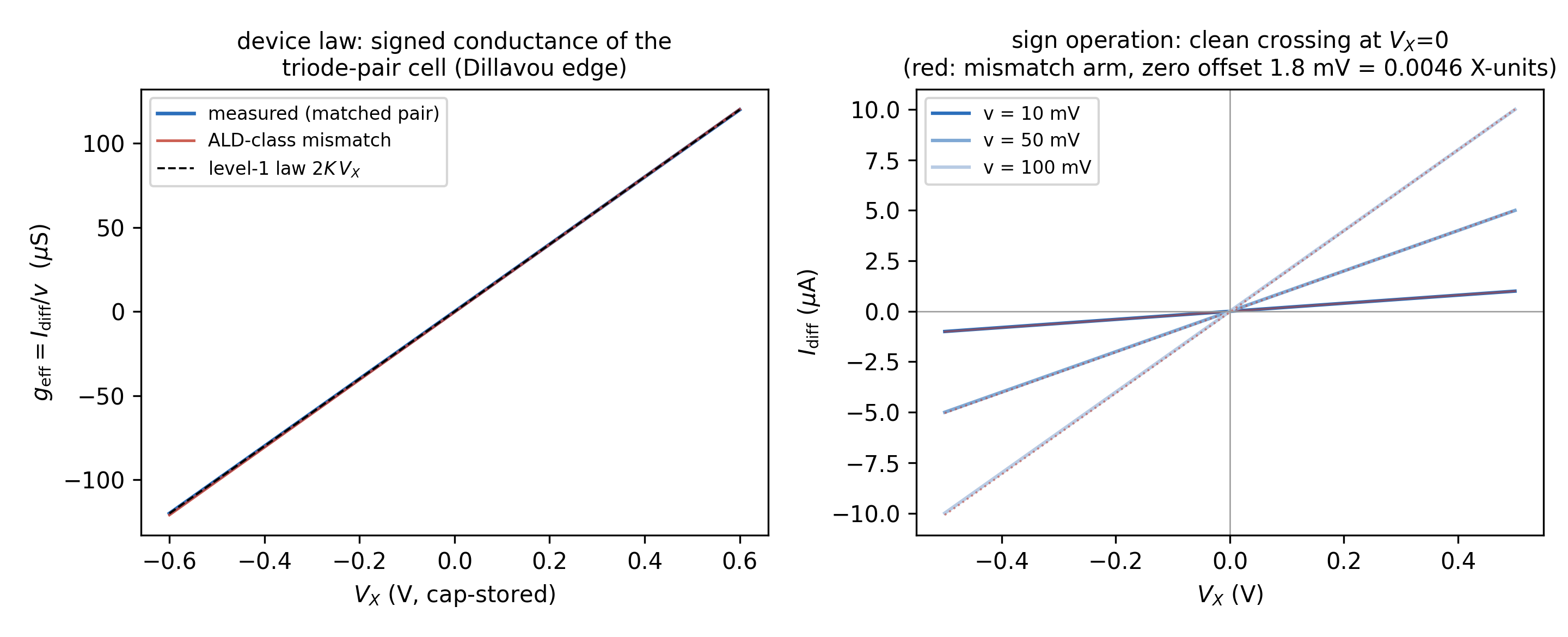}\hfill
\includegraphics[width=0.48\linewidth]{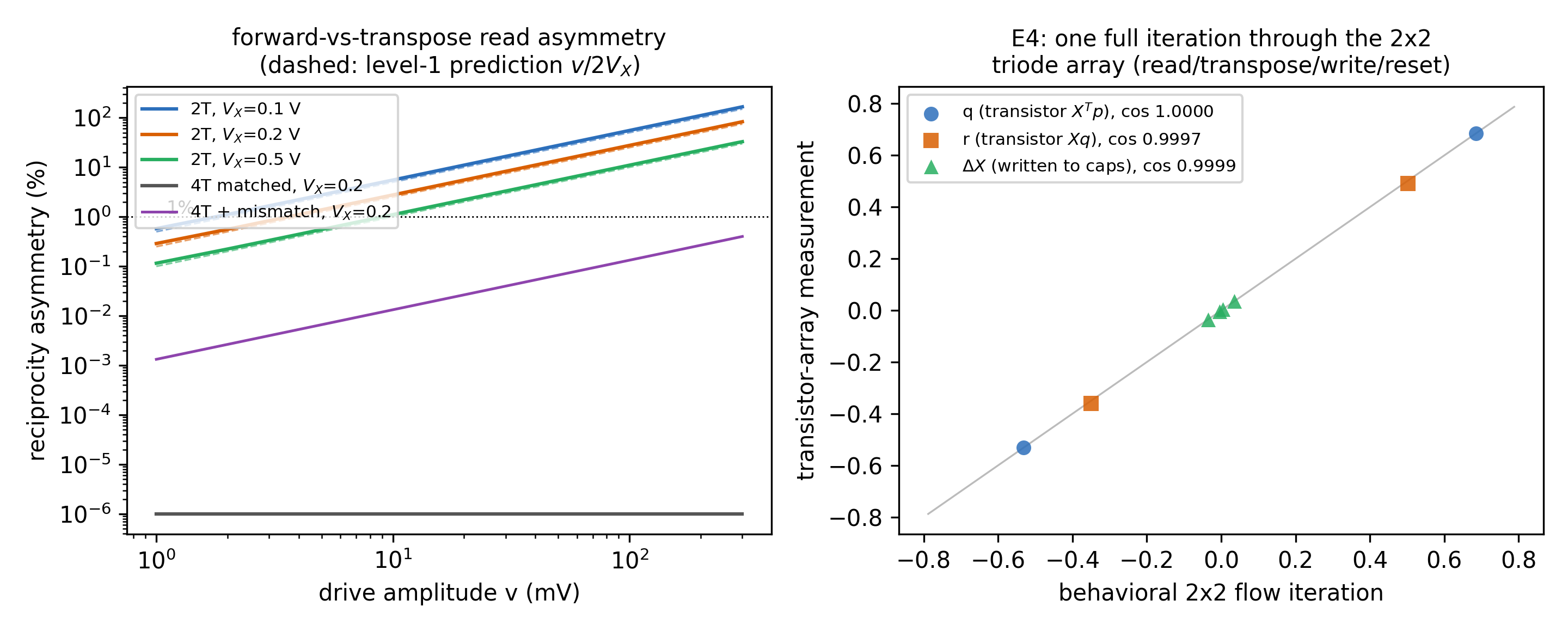}
\caption{Left: device law $g_{\mathrm{eff}} = 2KV_X$ of the read element (200.00 $\mu$S/V, deviation below 0.01\%). Right: reciprocity of the two- and four-transistor cells and the $2 \times 2$ array cosines for $q$, $r$, and $\Delta X$.}
\label{fig:cellC}
\end{figure}

\paragraph{Transistor-level read element.} Each entry of $X$ is read through two NMOS channels in the triode region with gates at $V_{\mathrm{ref}} \pm V_X$; the channels are the only elements in the signal path. The devices use a level-1 NMOS model with discrete-transistor parameters ($V_{T0} = 0.7$ V, $K_P = 0.1$ mA/V$^2$, $\gamma = 0.37$, $W/L = 100/100$ $\mu$m); a mismatch arm adds 2\% transconductance and 5 mV threshold mismatch. The device law $g_{\mathrm{eff}}(V_X) = 2K V_X$ holds with a simulated slope of 200.00 $\mu$S/V against a model value of 200 and a deviation from linearity below 0.01\% for $|V_X| \le 0.5$ V; the body-effect terms of the two channels cancel in pairs (Figure~\ref{fig:cellC}, left). The channel law $I = K(V_A - V_B)[(V_G - V_T) - (V_A + V_B)/2]$ is symmetric in its two terminals, so the device itself is reciprocal and the measured asymmetries are operating-point effects of the composite cell. The two-transistor cell reads exactly in the virtual-ground direction and shows a reverse asymmetry of $u / (2V_X)$ times a body-effect factor of 1.10 (measured 1.10--1.11); the sub-1\% region is $u \lesssim V_X / 55$ (3 mV at $V_X = 0.2$ V). The row-common part of the transpose residual is removed by the probe chopping, so the element-wise transpose error of 2.1\% leaves the cosine of $r$ at 0.9997. The four-transistor cell reads exactly in both directions with matched devices and shows 0.00\% small-signal and 0.13\% full-swing asymmetry under mismatch; the zero crossing is clean, with a mismatch offset of 1.8 mV (0.0046 in $X$ units). A full flow iteration on a $2 \times 2$ array with transistor reads gives cosines of 1.00000, 0.99972, and 0.99987 for $q$, $r$, and $\Delta X$; the write lands within $10^{-5}$ $X$ units, the read droop is 0.200 $\mu$V per 10 $\mu$s, the reset injection is 1.7 mV per capacitor, and the differential residual after reset is 0.00061 (Figure~\ref{fig:cellC}, right). The model boundaries are the level-1 device model (no mobility degradation or short-channel effects, adequate for discrete devices of this size), the $2 \times 2$ scale of the transistor-level array, and a behavioral DAC abstraction for the write chain.

\paragraph{Closed loop.} A four-class synthetic task trained with batch size 8 for 300 steps compares three orthogonalizers in the loop: digital NS5, the probe form with the behavioral array, and the probe form with the ngspice circuit. The three reach CE $< 0.01$ in 43, 42, and 42 steps and end at CE 0.0000 with 100\% training accuracy; the in-loop cosine of the orthogonalizer output against NS5 averages 0.895 for the behavioral model and 0.891 for the circuit (Figure~\ref{fig:toy}).

\end{document}